%% file: arxiv.tex
\documentclass[10pt,twocolumn,letterpaper]{article}

\makeatletter
\def\@fnsymbol#1{\ensuremath{\ifcase#1\or \dagger\or \ddagger\or
   \mathsection\or \mathparagraph\or \|\or **\or \dagger\dagger
   \or \ddagger\ddagger \else\@ctrerr\fi}}
\makeatother

\usepackage[pagenumbers]{iccv} 


\definecolor{iccvblue}{rgb}{0.21,0.49,0.74}
\usepackage[pagebackref,breaklinks,colorlinks,allcolors=iccvblue]{hyperref}
\usepackage{algorithm}
\usepackage{algorithmic}
\usepackage{amsmath}
\usepackage{booktabs}
\usepackage{multirow}
\usepackage{amssymb}
\usepackage[accsupp]{axessibility}
\newcommand{\B}[1]{\textbf{#1}}
\newcommand{\U}[1]{\underline{#1}}
\newcommand{\I}[1]{\textit{#1}}

\newcommand{\m}{ (ours)}

\def\paperID{5061} 
\def\confName{ICCV}
\def\confYear{2025}

\title{Knowledge Distillation with Refined Logits}

\author{Wujie Sun$^{1,2,3}$ \quad Defang Chen$^{4}$\thanks{Corresponding Author} \quad Siwei Lyu$^{4}$ \quad Genlang Chen$^{5}$ \quad Chun Chen$^{1,3}$ \quad Can Wang$^{1,3}$\\
$^{1}$State Key Laboratory of Blockchain and Data Security, Zhejiang University\\
$^{2}$School of Software Technology, Zhejiang University\\
$^{3}$Hangzhou High-Tech Zone (Binjiang) Institute of Blockchain and Data Security\\
$^{4}$University at Buffalo, State University of New York \quad $^{5}$NingboTech University\\
{\tt\small sunwujie@zju.edu.cn, \{defangch, siweilyu\}@buffalo.edu, \{cgl, chenc, wcan\}@zju.edu.cn}
}

\begin{document}
\maketitle
\begin{abstract}
    Recent research on knowledge distillation has increasingly focused on logit distillation because of its simplicity, effectiveness, and versatility in model compression. In this paper, we introduce Refined Logit Distillation (RLD) to address the limitations of current logit distillation methods. Our approach is motivated by the observation that even high-performing teacher models can make incorrect predictions, creating an exacerbated divergence between the standard distillation loss and the cross-entropy loss, which can undermine the consistency of the student model's learning objectives.
    Previous attempts to use labels to empirically correct teacher predictions may undermine the class correlations. In contrast, our RLD employs labeling information to dynamically refine teacher logits. In this way, our method can effectively eliminate misleading information from the teacher while preserving crucial class correlations, thus enhancing the value and efficiency of distilled knowledge. Experimental results on CIFAR-100 and ImageNet demonstrate its superiority over existing methods. 
    Our code is available at \url{https://github.com/zju-SWJ/RLD}.
\end{abstract}

\section{Introduction}

    Knowledge distillation utilizes pre-trained high-performing teacher models to facilitate the training of a compact student model~\cite{hinton2015distilling}. Compared to other model compression methods, such as pruning and quantization~\cite{choudhary2020comprehensive}, knowledge distillation exhibits fewer constraints on the model architecture. This flexibility significantly broadens its applicability, contributing to its increasing prominence in recent research.

    Hinton \etal~\cite{hinton2015distilling} were the first to introduce the concept of logit distillation. It is designed to align the logits of teacher and student models following the softmax operations with the Kullback-Leibler (KL) divergence. Most subsequent research has maintained the original concept of logit distillation, instead focusing on exploring feature distillation~\cite{romero2014fitnets,komodakis2017paying,tian2020contrastive,chen2021distilling,chen2022knowledge,wang2022semckd} in more depth by selecting and aligning intermediate-level features between teacher and student models. However, the potential architectural disparity between teacher and student models poses a significant challenge for feature alignment. This is mainly due to the fact that different architectures extract different features~\cite{wang2022semckd}. Moreover, the extensive diversity in feature selection further amplifies the complexity of feature distillation and leads to an increase in training time in distillation~\cite{chen2022knowledge}. Recently, by decoupling the classical logit distillation loss, Zhao \etal~\cite{zhao2022decoupled} demonstrate that logit distillation can yield results that are on par with, or even superior to, those of feature distillation. Consequently, logit distillation garnered considerable attention in the research community, thanks to its simplicity, effectiveness, and versatility.

\begin{figure}[t]
        \centering
        \includegraphics[width=\linewidth]{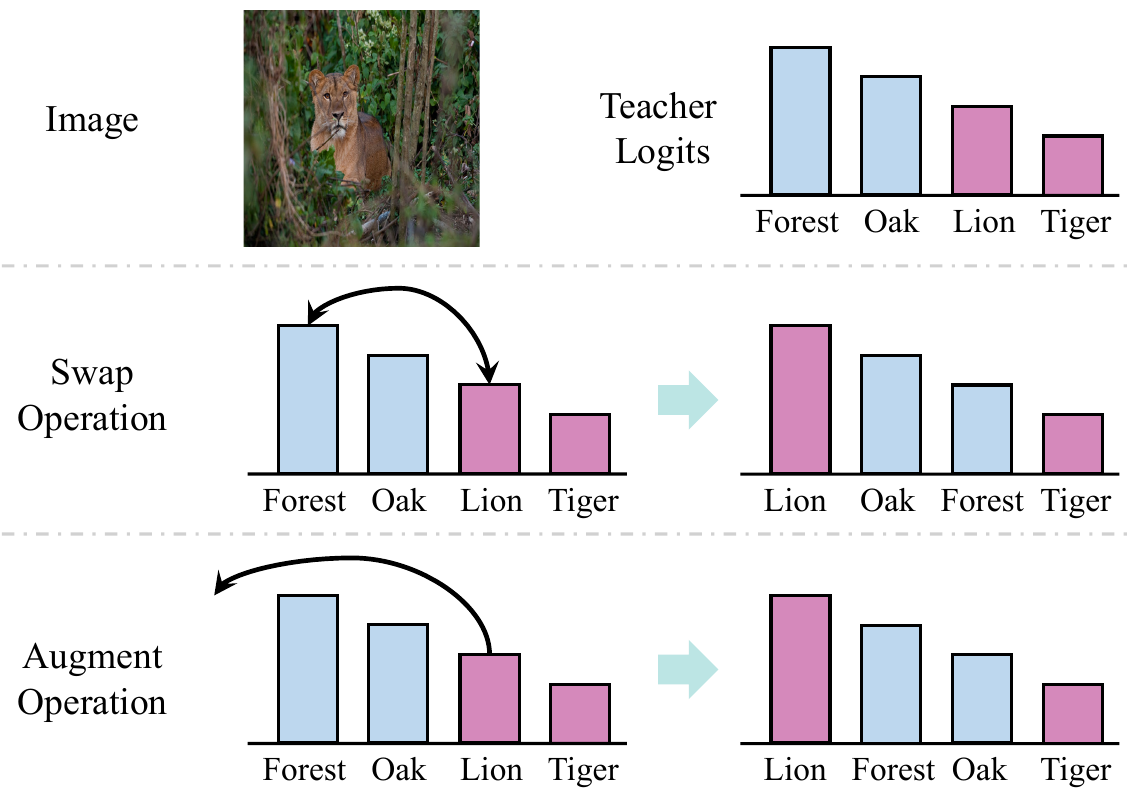}
        \caption{A toy example of existing correction-based distillation approaches. \textbf{Classes represented by the same color are highly correlated and should be ranked closely.} The image displayed is a ``lion'', yet the teacher model incorrectly classifies it as the ``forest''. Both the swap and augment operations disrupt the close correlation between ``lion'' and ``tiger''. A more detailed example of class correlation is provided in the Appendix.}
        \label{fig:correction}
    \end{figure}

    Despite the impressive achievements,  most of the recent logit distillation approaches~\cite{li2023curriculum,jin2023multi,sun2024logit} overlook the impact of teacher prediction correctness on the training process. Specifically, incorrect teacher predictions lead to an exacerbated divergence between teacher loss and label loss, which may severely impede the potential enhancements of the student models.
    Existing correction-based distillation approaches~\cite{wen2021preparing,cao2023excellent,lan2024improve} consistently modify the teacher logits (target) using label information. They either exchange the values between the predicted maximum class and the true class~\cite{wen2021preparing} (the {\em swap} operation) or amplify the proportion of the true class within the predicted probabilities~\cite{cao2023excellent,lan2024improve} (the {\em augment} operation). We argue that such approaches may alter the correlations among classes, as exemplified in \Cref{fig:correction}. This disruption can obstruct the transmission of ``dark knowledge''~\cite{hinton2015distilling} and hinder performance improvements.

    In this paper, we introduce {\em Refined Logit Distillation} (RLD) to address these challenges. 
    In classification tasks, the true class probability dictates prediction correctness, while class correlations capture high-level semantic relationships that influence classification tendencies. Accordingly, RLD consists of two types of knowledge, \textit{sample confidence} (SC) and \textit{masked correlation} (MC).
    Sample confidence refers to the binary probabilities derived from logits. As for the teacher model, SC comes from the probability associated with the predicted class and the probabilities of the remaining classes. It encapsulates the teacher's prediction confidence for the current sample and is employed to guide the student model. Considering the possible inaccuracies in the teacher's prediction, we align the student's true class probability with teacher's predicted class probability. This alignment not only mitigates the teacher's mistakes, but also guides the student model toward achieving a comparable level of confidence for the current sample. Moreover, it effectively prevents over-fitting.
    Masked correlation denotes our dynamic approach for selecting a subset of classes for teacher-student alignment. It is designed to mitigate the influence of potentially incorrect teacher predictions on student models while conveying essential class correlations. More specifically, MC involves masking all classes within the teacher logits that have equal or superior rankings compared to the true class. In essence, fewer classes are used for distillation when the teacher makes more mistakes, and more classes are used when it makes fewer mistakes. Using these two complementary types of refined knowledge, the student can achieve better performance.

    Our contributions are summarized as follows: \begin{itemize}
        \item We reveal that prevalent distillation approaches fail to account for the effects of incorrect teacher predictions, and existing correction-based strategies tend to ruin the valuable class correlations.
        \item We introduce a novel logit distillation approach termed Refined Logit Distillation (RLD) to prevent over-fitting and mitigate the influence of incorrect teacher knowledge, while preserving the essential class correlations.
        \item We conduct comprehensive experiments on CIFAR-100 and ImageNet datasets to verify the superior performance of our proposed RLD method.
    \end{itemize}
    
    \begin{figure*}[t]
        \centering
        \includegraphics[width=\linewidth]{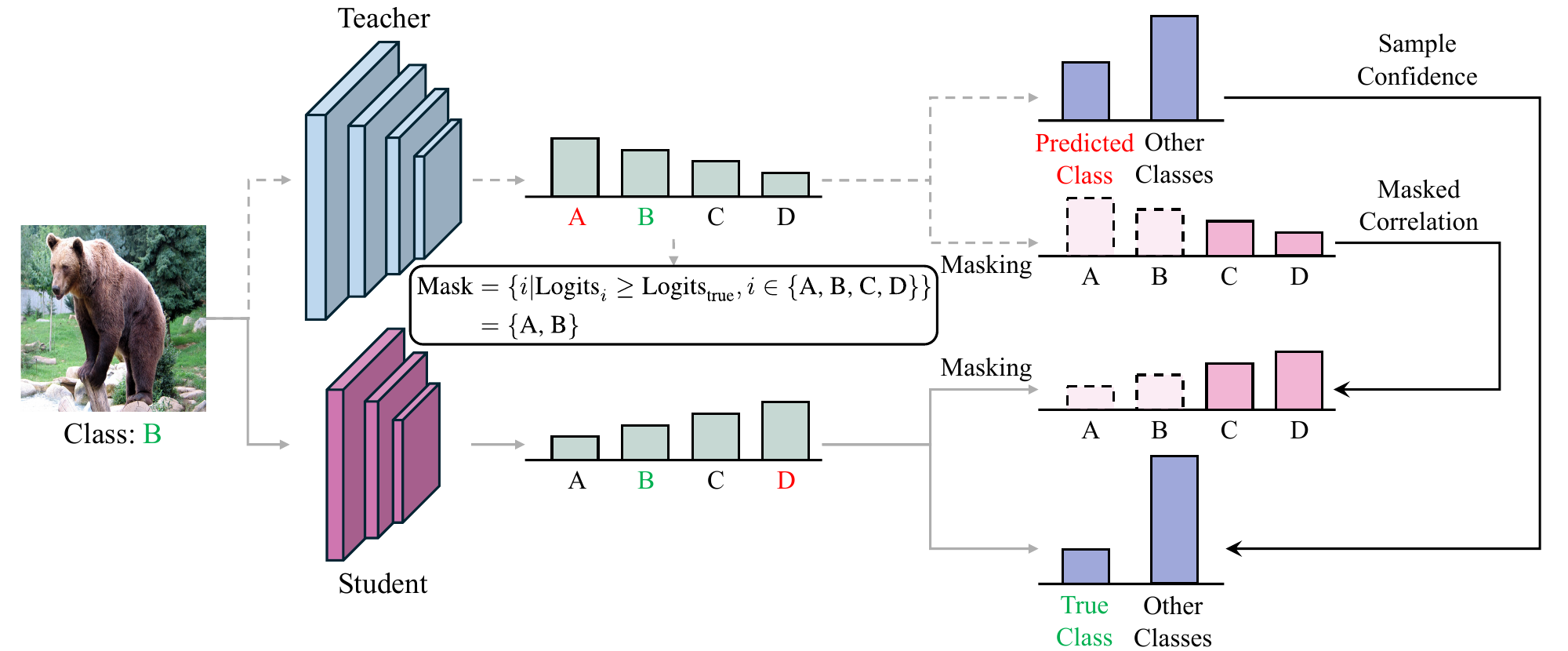}
        \caption{An overview of our proposed Refined Logit Distillation (RLD). In RLD, the teacher model imparts two types of knowledge, denoted as ``sample confidence'' and ``masked correlation'', to the student model. The binary sample confidence encapsulates the model confidence for each sample, which helps the student model generate proper-confidence predictions for the true class. The masked correlation denotes the probability distribution acquired after dynamically masking certain classes, which helps to remove misleading information and preserve valuable class correlations during distillation. Both kinds of knowledge are obtained from logits, thus the distillation process does not introduce intermediate layer features.}
        \label{fig:overview}
    \end{figure*}
    
\section{Related Work}

    The application of knowledge distillation historically concentrated on the image classification task, and progressively extended to a wider range of tasks, including semantic segmentation~\cite{liu2020structured,shu2021channel,yang2022cross,sun2023holistic} and image generation~\cite{salimans2022progressive,meng2023distillation,sun2023accelerating} within the realm of computer vision. Traditional knowledge distillation typically involves a single teacher and a single student model. As the field evolves, a variety of other paradigms have been proposed, such as online distillation~\cite{chen2020online,wu2021peer}, multi-teacher distillation~\cite{yuan2021reinforced,zhang2022confidence}, and self-distillation~\cite{furlanello2018born,kim2021self}. Since traditional knowledge distillation remains the core foundation of research in this area, we will focus solely on such methods in the discussion below.
    
    In image classification task, existing algorithms can be broadly classified into three categories: logit distillation~\cite{hinton2015distilling,zhao2022decoupled,jin2023multi,li2023curriculum,sun2024logit}, feature distillation~\cite{komodakis2017paying,kim2018paraphrasing,heo2019knowledge,tian2020contrastive,chen2021distilling,wang2022semckd,chen2022knowledge}, and relation distillation~\cite{park2019relational,liu2019knowledge,peng2019correlation}. Logit distillation has become the main focus of current research because of its straightforwardness, effectiveness, and adaptability. The initial logit distillation~\cite{hinton2015distilling} leverages KL divergence to align the softened output logits of the teacher and student models, thereby significantly enhancing the performance of the student models. DKD~\cite{zhao2022decoupled} revitalizes logit distillation by decoupling this classical loss, enabling it to perform comparably to feature distillation. MLKD~\cite{jin2023multi} leverages multi-level logit knowledge to further enhance model performance. CTKD~\cite{li2023curriculum} introduces the curriculum temperature, applying adversarial training and curriculum learning to dynamically determine the distillation temperature for each sample. LSKD~\cite{sun2024logit} processes the logits to adaptively allocate temperatures between teacher and student and across samples, thereby achieving state-of-the-art performance. However, the effect of incorrect teacher predictions on distillation is rarely considered. 
    
    Given that logits are intrinsically related to prediction correctness, several methods leverage labels to adjust logits prior to the distillation process. LA~\cite{wen2021preparing} swaps the values of the true and predicted classes to correct the teacher model's predictions. RC~\cite{cao2023excellent} adds the maximum value in the student's output to the true class, thereby aiding the student model in making accurate and confident predictions. LR~\cite{lan2024improve} combines one-hot labels with the teacher's soft labels to produce a new, precise target for distillation. However, as previously demonstrated in \Cref{fig:correction}, these methods may disrupt class correlations, which can hinder performance improvement.
    
\section{Preliminaries}

    We provide an overview of concepts related to knowledge distillation to facilitate readers' understanding.
    
    Consider an image classification task involving $C$ classes. We have a pre-trained teacher model and a student model, denoted as $\theta^\text{T}$ and $\theta^\text{S}$, respectively. For a single input image $x$, the output logits $z$ from teacher and student models are denoted as $z^\text{T}=\theta^\text{T}\left(x\right)$ and $z^\text{S}=\theta^\text{S}\left(x\right)$, respectively. By utilizing the softmax function $\sigma\left(\cdot\right)$, predicted distributions $p^\text{T}$ and $p^\text{S}$ are calculated as follows:
    \begin{equation}\label{eq:softmax}
        p_i
        =\frac{\exp\left({z_i}\right)}{\sum^{C}_{c=1}\exp\left({z_c}\right)},
    \end{equation}
    where $p_i$ represents the predicted value of the $i$-th class. 
    
    To train the student model, the first loss is computed as the cross entropy between the student prediction and the one-hot ground-truth label $y$:
    \begin{equation}\label{eq:ce}
        L_{\text{CE}}
        =-\sum^{C}_{c=1}y_c\log p^\text{S}_c.
    \end{equation}
    
    The second loss aligns the softened predictions $\hat{p}=\sigma\left(z/\tau\right)$ of the teacher and student models using the KL divergence:
    \begin{equation}\label{eq:kd}
        L_{\text{KD}}
        =\tau^2\text{KL}\left(\hat{p}^\text{T},\hat{p}^\text{S}\right)
        =\tau^2\sum^{C}_{c=1}\hat{p}_c^\text{T}\log\frac{\hat{p}_c^\text{T}}{\hat{p}_c^\text{S}},
    \end{equation}
    where $\tau$ denotes the temperature for the softmax operation.

    By combining \Cref{eq:ce,eq:kd}, we get the classical logit distillation loss for stochastic gradient descent. Such an approach has been experimentally shown to perform better than training solely with labels.
    
\section{Methodology}

    In this section, we delve into a detailed introduction of our proposed RLD. An overview of RLD is shown in \Cref{fig:overview}. 
    
    \subsection{Sample Confidence Distillation}
    
        Sample confidence (SC) represents the binary distribution $b$ derived from the logits. It encapsulates the model confidence for each sample, thereby aiding the student model in generating proper-confidence predictions for the true class, without unduly restricting the distribution for other classes.

        In the context of teacher knowledge, one component of the SC is the maximum predicted probability value $\hat{p}_\text{max}^\text{T}$, while the other component is the sum of the predicted probabilities for the remaining classes. In contrast, the student SC consists of two components: the predicted probability for the true class $\hat{p}_\text{true}^\text{S}$, and the sum of the predicted probabilities for the remaining classes. They can be summarized in the following formulas:
        \begin{equation}\label{eq:binary_t}
            b^\text{T}
            =\{\hat{p}_\text{max}^\text{T},1-\hat{p}_\text{max}^\text{T}\},
        \end{equation}
        \begin{equation}\label{eq:binary_s}
            b^\text{S}
            =\{\hat{p}_\text{true}^\text{S},1-\hat{p}_\text{true}^\text{S}\}.
        \end{equation}

        To transfer this knowledge, we align $b^\text{T}$ and $b^\text{S}$ using the KL divergence:
        \begin{equation}\label{eq:scd}
            L_{\text{SCD}}
            =\tau^2\text{KL}\left(b^\text{T},b^\text{S}\right).
        \end{equation}

\begin{figure}[t]
            \centering
            \includegraphics[width=\linewidth]{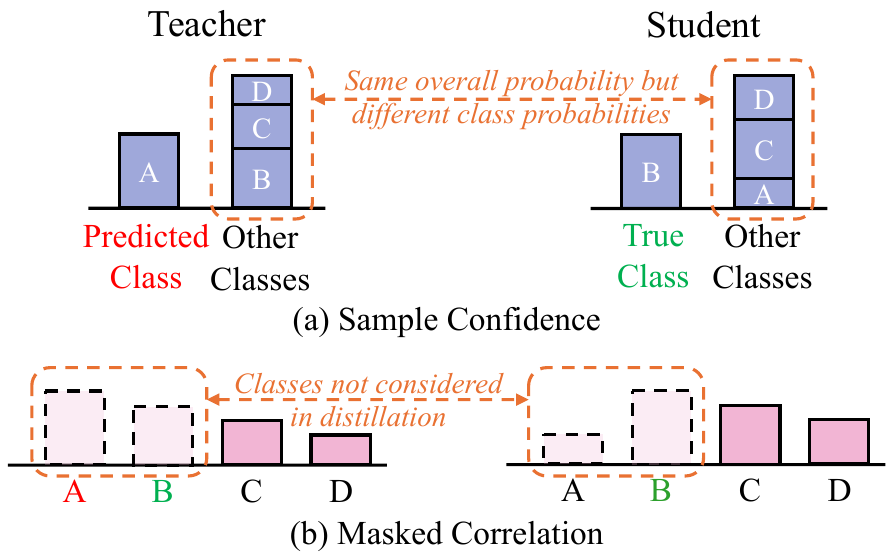}
            \caption{Toy examples elucidating the variances across methods \textbf{when the corresponding losses are minimal}. (a) Sample confidence guarantees a proper probability for the true class and eliminates the necessity for the student model to match the intact class probability distribution to that of the teacher model. (b) Masked correlation imposes fewer constraints on the student prediction than traditional knowledge distillation. For the classes that are masked, their probabilities can diverge entirely from those of the teacher model.}
            \label{fig:example}
        \end{figure}

        \Cref{fig:example}(a) more vividly illustrates the aligned knowledge of the teacher and the student when using SCD.
        
        While both $L_{\text{CE}}$ and $L_{\text{SCD}}$ operate on the true class, gradient analysis (temperature $\tau$ omitted) shows that their effects are not entirely identical:
        \begin{equation}
        \begin{gathered}
            \frac{\partial L_{\mathrm{CE}}}{\partial z_i} = p^{\text{S}}_i-y_i, \\
            \frac{\partial L_{\mathrm{SCD}}}{\partial z_{i}} =
                \begin{cases}
            		p^{\text{S}}_i-p^{\text{T}}_{\text{max}},
            		& \mbox{$i=\text{true}$}, \\
            		\frac{p^{\text{S}}_i(p^{\text{S}}_{\text{true}}-p^{\text{T}}_{\text{max}})}{p^{\text{S}}_{\text{true}}-1},	  
            		& \mbox{$i\neq\text{true}$}.
                \end{cases}
        \end{gathered}
        \end{equation}
        Consequently, RLD incorporates both, thereby endowing the distillation process with greater flexibility.
        
    \subsection{Masked Correlation Distillation}
        Masked correlation (MC) denotes the probability distribution acquired after dynamically masking certain classes. As shown in \Cref{fig:example}(b), this masking operation relieves the student model from aligning incorrect class rankings, thereby allowing the student model to generate very different output from the teacher without incurring a large loss. Moreover, preserving partial class probabilities empowers the student model to learn valuable class correlations, consequently enhancing the model's performance.
        
        Specifically, the mask $M$ is dynamically derived from teacher logits and labels. We designate all classes whose logit values are \textbf{g}reater than or \textbf{e}qual to (denoted as ``ge'') the logit value of the true class as the targets for the masking operation, which can be represented as follows:
        \begin{equation}\label{eq:mask}
            M_\text{ge}
            =\{i|z_i^\text{T}\ge z_\text{true}^\text{T},1\le i \le C\}.
        \end{equation}

        After obtaining the mask, we compute the probability distributions for alignment using the following formula:
        \begin{equation}\label{eq:masked_p}
            \tilde{p}_i
            =\frac{\exp\left({z_i/\tau}\right)}{\sum^{C}_{c=1,c\not\in M_\text{ge}}\exp\left({z_c/\tau}\right)},
        \end{equation}
        where $1\le i \le C$ and $i\not\in M_\text{ge}$ are satisfied.
    
        We summarize the distillation loss for the masked correlation knowledge as follows:
        \begin{equation}\label{eq:mcd}
            L_{\text{MCD}}
            =\tau^2\text{KL}\left(\tilde{p}^\text{T},\tilde{p}^\text{S}\right).
        \end{equation}
        
        When the teacher model makes a more accurate prediction (ranking the true class higher), only a few classes are subjected to the masking operation. It allows the majority of class correlations to be preserved and transferred to the student model. Conversely, if the teacher's prediction is less accurate, the majority of classes are masked. As a result, the student model learns less knowledge, thereby reducing the potential for misinformation to mislead the training process. It also gives the student model more freedom to make predictions for masked classes that differ significantly from those of the teacher model.

    \subsection{Refined Logit Distillation}
        
        By combining \Cref{eq:ce,eq:scd,eq:mcd}, we obtain the final loss for RLD, which is:
        \begin{equation}\label{eq:RLD}
            L_{\text{RLD}}
            =L_{\text{CE}}+\alpha L_{\text{SCD}}+\beta L_{\text{MCD}},
        \end{equation}        
        where hyper-parameters $\alpha$ and $\beta$ adjust the importance of sample confidence and masked correlation, respectively.

        \begin{table*}[!t]
            \centering
            \begin{tabular}{llccccccc}
            \toprule
            \multirow{4}{*}{Type} 
            & \multirow{2}{*}{Teacher} 
            & ResNet32$\times$4 & VGG13 & WRN-40-2 & ResNet56 & ResNet110 & ResNet110 \\
            &          &   79.42 &   74.64 &   75.61 &   72.34 &   74.31 &   74.31 \\
            & \multirow{2}{*}{Student} 
            & ResNet8$\times$4 & VGG8 & WRN-40-1 & ResNet20 & ResNet32 & ResNet20 \\
            &          &   72.50 &   70.36 &   71.98 &   69.06 &   71.14 &   69.06 \\
            \midrule
            \multirow{8}{*}{Feature} 
            & FitNet   &   73.50 &   71.02 &   72.24 &   69.21 &   71.06 &   68.99 \\
            & AT       &   73.44 &   71.43 &   72.77 &   70.55 &   72.31 &   70.65 \\
            & RKD      &   71.90 &   71.48 &   72.22 &   69.61 &   71.82 &   69.25 \\
            & CRD      &   75.51 &   73.94 &   74.14 &   71.16 &   73.48 &   71.46 \\
            & OFD      &   74.95 &   73.95 &   74.33 &   70.98 &   73.23 &   71.29 \\
            & ReviewKD &   75.63 &   74.84 &\I{75.09}&   71.89 &   73.89 &   71.34 \\
            & SimKD    &\I{78.08}&   74.89 &   74.53 &   71.05 &   73.92 &   71.06 \\
            & CAT-KD   &   76.91 &   74.65 &   74.82 &   71.62 &   73.62 &   71.37 \\
            \midrule
            \multirow{7}{*}{Logit} 
            & KD       &   73.33 &   72.98 &   73.54 &   70.66 &   73.08 &   70.67 \\
            & CTKD     &   73.39 &   73.52 &   73.93 &   71.19 &   73.52 &   70.99 \\
            & DKD      &\U{76.32}&\U{74.68}&\U{74.81}&\U{71.97}&\B{74.11}&   71.06 \\
            & LA       &   73.46 &   73.51 &   73.75 &   71.24 &   73.39 &   70.86 \\
            & RC       &   74.68 &   73.37 &   74.07 &   71.63 &   73.44 &\U{71.41}\\
            & LR       &   76.06 &   74.66 &   74.42 &   70.74 &   73.52 &   70.61 \\
            & RLD\m    &\B{76.64}&\B{74.93}&\B{74.88}&\B{72.00}&\U{74.02}&\B{71.67}\\
            \bottomrule
            \end{tabular}
            \caption{Top-1 accuracy (\%) on the CIFAR-100 validation set when the teacher and student models are homogeneous. The best and second best results of logit distillation are highlighted in \textbf{bold} and \underline{underlined} text, respectively. For the case where the best result of feature distillation is better than the best result of logit distillation, we highlight it with \textit{italic} text. The reported results are the mean of three trials.}
            \label{tab:sameStru}
        \end{table*}
        
        \paragraph{Relevance to DKD.} Although RLD and DKD~\cite{zhao2022decoupled} consider logit distillation from distinct perspectives, they become equivalent when the teacher model consistently makes accurate predictions. Besides, DKD does not explicitly explain why transferring non-target class knowledge (i.e., the probability distribution when the true class is masked, referred to as NCKD) can significantly enhance model performance. Beyond the idea that the class relationships embedded in this knowledge facilitate training, RLD offers a new explanation: when the alignment constraint on target class knowledge (TCKD) is weak, masking the true class during distribution alignment offers the student model greater flexibility to adjust the ranking of the true class. This, in turn, mitigates the negative impact of incorrect teacher knowledge, enabling more accurate predictions.
        
\section{Experiments}
        
    \subsection{Settings}
    
        \begin{table*}[!t]
            \centering
            \begin{tabular}{llccccccc}
            \toprule
            \multirow{4}{*}{Type} 
            & \multirow{2}{*}{Teacher} 
            & ResNet32$\times$4 & ResNet32$\times$4 & WRN-40-2 & WRN-40-2 & VGG13 & ResNet50 \\
            &          &   79.42 &   79.42 &   75.61 &   75.61 &   74.64 &   79.34 \\
            & \multirow{2}{*}{Student} 
            & SHN-V2 & WRN-40-2 & ResNet8$\times$4 & MN-V2 & MN-V2 & MN-V2 \\
            &          &   71.82 &   75.61 &   72.50 &   64.60 &   64.60 &   64.60 \\
            \midrule
            \multirow{8}{*}{Feature} 
            & FitNet   &   73.54 &   77.69 &   74.61 &   68.64 &   64.16 &   63.16 \\
            & AT       &   72.73 &   77.43 &   74.11 &   60.78 &   59.40 &   58.58 \\
            & RKD      &   73.21 &   77.82 &   75.26 &   69.27 &   64.52 &   64.43 \\
            & CRD      &   75.65 &   78.15 &   75.24 &   70.28 &   69.73 &   69.11 \\
            & OFD      &   76.82 &   79.25 &   74.36 &   69.92 &   69.48 &   69.04 \\
            & ReviewKD &   77.78 &   78.96 &   74.34 &\I{71.28}&\I{70.37}&   69.89 \\
            & SimKD    &   78.39 &\I{79.29}&   75.29 &   70.10 &   69.44 &   69.97 \\
            & CAT-KD   &\I{78.41}&   78.59 &   75.38 &   70.24 &   69.13 &\I{71.36}\\
            \midrule
            \multirow{7}{*}{Logit} 
            & KD       &   74.45 &   77.70 &   73.97 &   68.36 &   67.37 &   67.35 \\
            & CTKD     &   75.37 &   77.66 &   74.61 &   68.34 &   68.50 &   68.67 \\
            & DKD      &\U{77.07}&   78.46 &\U{75.56}&\U{69.28}&   69.71 &   70.35 \\
            & LA       &   75.14 &   77.39 &   73.88 &   68.57 &   68.09 &   68.85 \\
            & RC       &   75.61 &   77.58 &   75.22 &   68.72 &   68.66 &   68.98 \\
            & LR       &   76.27 &\U{78.73}&   75.26 &   69.02 &\U{69.78}&\U{70.38}\\
            & RLD\m    &\B{77.56}&\B{78.91}&\B{76.12}&\B{69.75}&\B{69.97}&\B{70.76}\\
            \bottomrule
            \end{tabular}
            \caption{Top-1 accuracy (\%) on the CIFAR-100 validation set when the teacher and student models are heterogeneous. The same convention is used as in \Cref{tab:sameStru}.}
            \label{tab:diffStru}
        \end{table*}

        \paragraph{Datasets.} We conduct the experiments on two standard image classification datasets: CIFAR-100~\cite{krizhevsky2009learning} and ImageNet~\cite{russakovsky2015imagenet}. CIFAR-100 comprises 100 distinct classes, with a total of 50,000 images in the training set and 10,000 images in the validation set. Each image in this dataset is of size 32$\times$32 pixels. ImageNet presents a larger and more complex dataset, encompassing 1,000 classes. It includes 1.28 million images in the training set and 50,000 images in the validation set, with each image resolution being 224$\times$224 pixels after pre-processing. 

        \paragraph{Models.} Models used by teachers and students include ResNet~\cite{he2016deep}, WideResNet (WRN)~\cite{zagoruyko2016wide}, VGG~\cite{simonyan2015very}, ShuffleNet (SHN)~\cite{ma2018shufflenet,zhang2018shufflenet}, and MobileNet (MN)~\cite{howard2017mobilenets,sandler2018mobilenetv2}. 

        \paragraph{Compared Methods.}\label{sec:compared_methods} We emphasize that the experimental results in this paper are presented after exhaustive reading of the papers and codes of existing works, and with a focus on fair comparisons between methods. Therefore, MLKD~\cite{jin2023multi} is excluded from the comparison due to differing experimental settings. The compared methods include feature distillation (FitNet~\cite{romero2014fitnets}, AT~\cite{komodakis2017paying}, RKD~\cite{park2019relational}, CRD~\cite{tian2020contrastive}, OFD~\cite{heo2019comprehensive}, ReviewKD~\cite{chen2021distilling}, SimKD~\cite{chen2022knowledge}, and CAT-KD~\cite{guo2023class}) and logit distillation (KD~\cite{hinton2015distilling}, CTKD~\cite{li2023curriculum}, DKD~\cite{zhao2022decoupled}, LA~\cite{wen2021preparing}, RC~\cite{cao2023excellent}, and LR~\cite{lan2024improve}) methods. The performance metrics of all comparison methods, except for the latter three correction-based approaches (LA, RC, and LR), are sourced directly from LSKD~\cite{sun2024logit}. To ensure experimental fairness, we implemented these three correction-based approaches and our proposed RLD method following the experimental setups commonly used in prominent studies (e.g., LSKD~\cite{sun2024logit}, DKD~\cite{zhao2022decoupled}, and CRD~\cite{tian2020contrastive}). Additional implementation details are provided in the Appendix.

    \subsection{Main Results}\label{sec:result}
        
        \paragraph{CIFAR-100.} The top-1 validation accuracy (\%) comparison results of RLD and other distillation approaches are reported in \Cref{tab:sameStru} (homogeneous distillation pairs) and \Cref{tab:diffStru} (heterogeneous distillation pairs). We can see that RLD is either the optimal or suboptimal logit distillation algorithm in all cases, and is optimal in most cases. This underscores the superiority of RLD and accentuates the significance of making corrections to teacher predictions. While feature distillation can sometimes outperform logit distillation, its optimal method varies across teacher-student pairs, and its longer training time and complex algorithm design may hinder practical applicability.
    
        \begin{table}[!t]
            \centering
            \begin{tabular}{lcccc}
            \toprule
            Teacher/Student
            & \multicolumn{2}{l}{Res34/Res18} & \multicolumn{2}{l}{Res50/MN-V1} \\
            \midrule
            Accuracy &   Top-1 &   Top-5 &   Top-1 &   Top-5 \\
            \midrule
            Teacher  &   73.31 &   91.42 &   76.16 &   92.86 \\
            Student  &   69.75 &   89.07 &   68.87 &   88.76 \\
            \midrule
            AT       &   70.69 &   90.01 &   69.56 &   89.33 \\
            OFD      &   70.81 &   89.98 &   71.25 &   90.34 \\
            CRD      &   71.17 &   90.13 &   71.37 &   90.41 \\
            ReviewKD &   71.61 &\U{90.51}&\U{72.56}&   91.00 \\
            SimKD    &   71.59 &   90.48 &   72.25 &   90.86 \\
            CAT-KD   &   71.26 &   90.45 &   72.24 &\U{91.13}\\
            \midrule
            KD       &   71.03 &   90.05 &   70.50 &   89.80 \\
            CTKD     &   71.38 &   90.27 &   71.16 &   90.11 \\
            DKD      &\U{71.70}&   90.41 &   72.05 &   91.05 \\
            LA       &   71.17 &   90.16 &   70.98 &   90.13 \\
            RC       &   71.59 &   90.21 &   71.86 &   90.54 \\
            LR       &   70.29 &   89.98 &   71.76 &   90.93 \\
            RLD\m    &\B{71.91}&\B{90.59}&\B{72.75}&\B{91.18}\\
            \bottomrule
            \end{tabular}
            \caption{Top-1 and top-5 accuracy (\%) on the ImageNet validation set. The best and second best results are highlighted in \textbf{bold} and \underline{underlined} text, respectively. The reported results are the mean of three trials.}
            \label{tab:imagenet}
        \end{table}

        \paragraph{ImageNet.} The top-1 and top-5 validation accuracy (\%) comparison results of RLD and other distillation approaches are reported in \Cref{tab:imagenet}. On this more challenging dataset, RLD successfully outperforms all existing feature and logit distillation algorithms, consistently achieving optimal performance and demonstrating its superiority.
        
\begin{figure}[t]
            \centering
            \includegraphics[width=\linewidth]{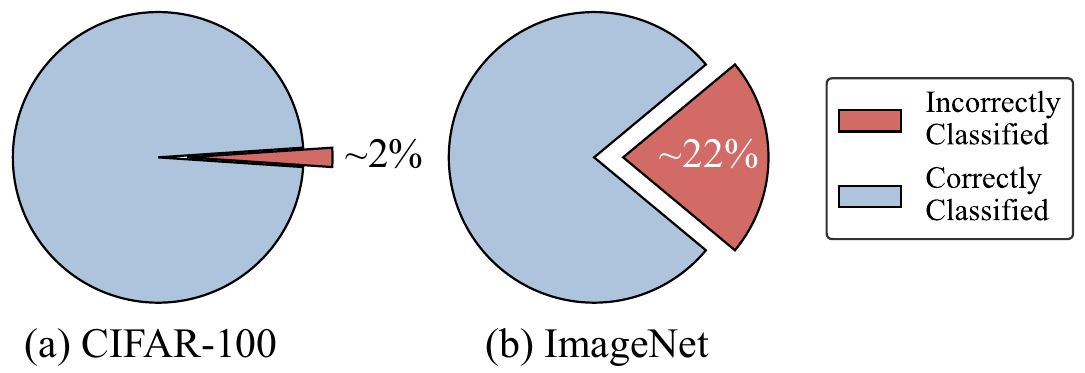}
            \caption{Proportion of predictions from teacher models on the training set. (a) ResNet56. (b) ResNet50.}
            \label{fig:proportion}
        \end{figure}

        \paragraph{Analysis.} Examining the experimental results detailed in \Cref{tab:diffStru,tab:sameStru,tab:imagenet}, it is clear that the performance improvement brought about by RLD is more substantial on the ImageNet dataset than on the CIFAR-100 dataset. We presume that this discrepancy stems from the varying accuracy levels of the teacher models on the respective training sets. As shown in \Cref{fig:proportion}, the teacher model exhibits high training accuracy on the CIFAR-100 dataset, while it shows comparatively lower accuracy on the ImageNet dataset. Given that RLD aligns with DKD when the teacher predictions are accurate, the high training accuracy on the CIFAR-100 dataset might hinder substantial divergence between these two methods. Conversely, on the ImageNet dataset, where the training accuracy is lower, RLD outperforms DKD by achieving a more substantial improvement.
        
    \subsection{Extensions}\label{sec:extensions}

        \begin{table}[!t]
            \centering
            \begin{tabular}{lccc}
            \toprule
            \multirow{2}{*}{Teacher} & ResNet56 & ResNet110&   VGG13  \\
                                     &   72.34  &   74.31  &   74.64  \\
            \midrule
            \multirow{2}{*}{Student} & \multicolumn{3}{c}{WideResNet-40-2}   \\
                                     & \multicolumn{3}{c}{75.61}      \\
            \midrule
            KD                       &   76.72  &   77.37  &   76.86  \\
            DKD                      &   77.34  &   77.70  &   77.45  \\
            RLD\m                    &\B{78.03} &\B{78.28} &\B{77.88} \\
            \midrule
            $\Delta$                 &   +0.69  &   +0.58  &   +0.43  \\
            \bottomrule
            \end{tabular}
            \caption{Top-1 accuracy (\%) on the CIFAR-100 validation set when distilling with inferior teachers. Optimal results are highlighted in \textbf{bold}. The reported results are the mean of three trials.}
            \label{tab:worse_teacher}
        \end{table}
        
        \paragraph{Reversed Knowledge Distillation.} We explore a unique scenario termed reversed knowledge distillation~\cite{yuan2020revisiting}, where the teacher performs worse than the student. This study investigates the feasibility of using an inferior teacher model to enhance the performance of a superior student model, particularly in situations where sourcing a more capable teacher model proves challenging. As shown in \Cref{tab:worse_teacher}, among all distillation pairs, RLD achieves the best performance, and shows a large performance gap compared to DKD. Moreover, the accuracy difference $\Delta$ between RLD and DKD shows that poorer teacher model performance leads to greater RLD improvement over DKD. This can be attributed to two main factors: firstly, the use of the inferior teacher allows RLD to be better distinguished from DKD; secondly, the unique setup of reversed knowledge distillation imposes more stringent demands on the quality of knowledge transferred, thereby underscoring the effectiveness of RLD in refining distilled knowledge.

        \begin{table}[!t]
            \centering
            \begin{tabular}{lccc}
            \toprule
            \multirow{2}{*}{Teacher} & WRN-40-2 &   VGG13  & ResNet50 \\
                                     & 75.61    &   74.64  & 79.34    \\
            \midrule
            \multirow{2}{*}{Student} & \multicolumn{3}{c}{MobileNet-V2} \\
                                     & \multicolumn{3}{c}{64.60}      \\
            \midrule
            KD                       &   69.23  &   68.61  &   69.02  \\
            CTKD                     &   69.53  &   68.98  &   69.36  \\
            DKD                      &   70.01  &   69.98  &   70.45  \\
            RLD\m                    &\B{70.35} &\B{70.63} &\B{71.06} \\
            \bottomrule
            \end{tabular}
            \caption{Top-1 accuracy (\%) on the CIFAR-100 validation set when training with logit standardization technique LSKD~\cite{sun2024logit}. The optimal results are highlighted in \textbf{bold} text. The reported results are the mean of three trials.}
            \label{tab:standardization}
        \end{table}

        \paragraph{Logit Standardization.} We investigate the efficacy of each method when supplemented with logit standardization technique LSKD~\cite{sun2024logit}. The results are shown in \Cref{tab:standardization}. The optimal results achieved by RLD underscore its superior performance and the vast potential of its integration with other methodologies.
        
        \begin{figure}[t]
            \centering
            \includegraphics[width=\linewidth]{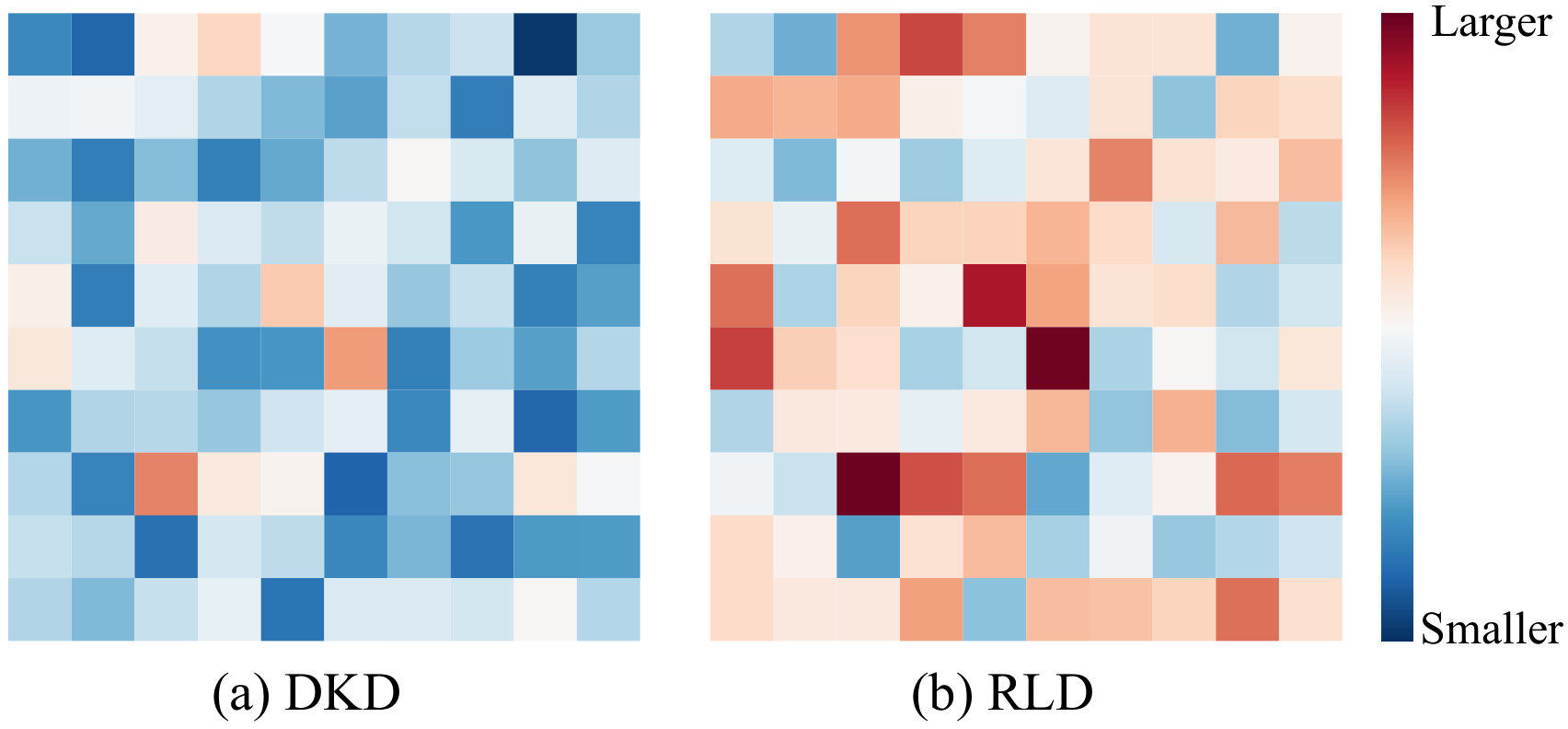}
            \caption{Visualized teacher-student logit discrepancy learned by DKD and RLD on the CIFAR-100 validation set. For better visualization, 100 classes are reshaped into a 10$\times$10 matrix. The teacher is ResNet32$\times$4, and the student is ResNet8$\times$4.}
            \label{fig:diff}
        \end{figure}
        
        \paragraph{Logit Discrepancy Visualization.} We calculate the mean absolute error (MAE) of logits for each class between teacher and student models obtained via DKD and RLD, visualizing these results using the heat map in \Cref{fig:diff}. Despite RLD outperforming DKD, it is observed that the logit discrepancy yielded by RLD is larger than that of DKD. This observation aligns with our anticipation, given that RLD rectifies certain inaccuracies in teacher knowledge and provides students with greater autonomy in formulating their own predictions. This finding underscores that an unconsidered alignment with teacher knowledge may not be the optimal strategy, and we believe that correction-based approaches deserve more attention and research.
        
\begin{figure}[t]
            \centering
            \includegraphics[width=\linewidth]{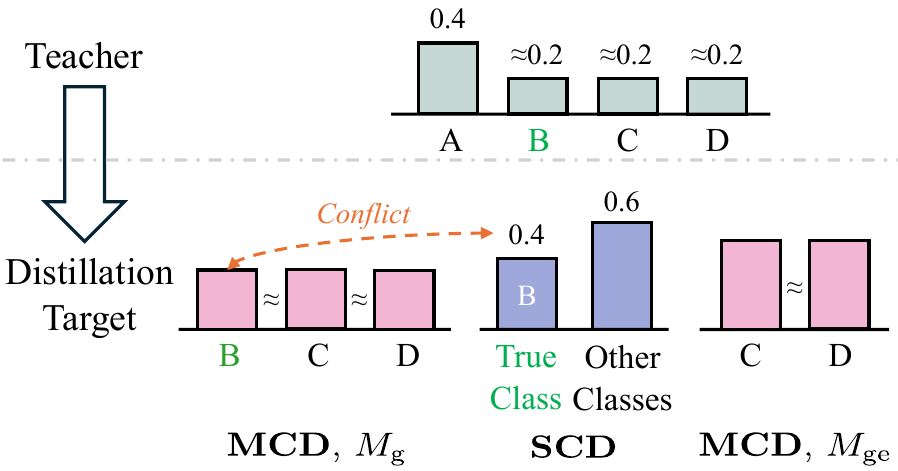}
            \caption{A toy example illustrates how the $M_\text{g}$ masking strategy can lead to loss conflict. In this example, the probabilities of classes B, C, and D in the teacher distribution are close, and the value corresponding to B is slightly larger than those of C and D.}
            \label{fig:masking}
        \end{figure}

        \paragraph{Ablation Study.} We perform an ablation study on the components of RLD, and the results are shown in \Cref{tab:component}. The results demonstrate that each component of RLD effectively contributes to enhanced performance. Notably, while masking all classes with values \textbf{g}reater than (denoted as ``g'') those of the true class would similarly eliminate misinformation and preserve class correlations, this masking strategy (denoted as $M_\text{g}$) inadvertently integrates true class-related knowledge into both masked correlation and sample confidence. This overlap may create conflicts between the resulting losses, thereby hindering performance improvement. \Cref{fig:masking} presents a toy example for illustration. After applying $M_\text{g}$, the distillation objective of MCD requires that the probabilities of classes B, C, and D be close. This conflicts with the objective of SCD, which enforces a probability of 0.4 for class B, since the sum of class probabilities cannot exceed 1. In contrast, this issue does not arise when using $M_\text{ge}$. Therefore, we opt for $M_\text{ge}$ as the masking strategy.

        \begin{table}[t]
            \centering
            \begin{tabular}{ccccc}
            \toprule
            \multirow{2}{*}{$L_\text{CE}$} & \multirow{2}{*}{$L_\text{SCD}$} & \multicolumn{2}{c}{$L_\text{MCD}$} & \multirow{2}{*}{Accuracy} \\
                                           &                                 & $M_\text{g}$ & $M_\text{ge}$       &                           \\
            \midrule
            \checkmark    &                &                &                & 72.50    \\
            \midrule
            \checkmark    & \checkmark     &                &                & 73.55    \\
            \midrule
            \checkmark    &                & \checkmark     &                & 75.50    \\
            \checkmark    &                &                & \checkmark     & 75.64    \\
            \midrule
            \checkmark    & \checkmark     & \checkmark     &                & 75.53    \\
            \checkmark    & \checkmark     &                & \checkmark     & \B{76.64}\\
            \bottomrule
            \end{tabular}
            \caption{Ablation study on the importance of each component in RLD. Top-1 accuracy (\%) on the CIFAR-100 validation set is reported. The teacher is ResNet32$\times$4, and the student is ResNet8$\times$4. The reported results are the mean of three trials.}
            \label{tab:component}
        \end{table}

        \begin{figure}[t]
            \centering
            \includegraphics[width=\linewidth]{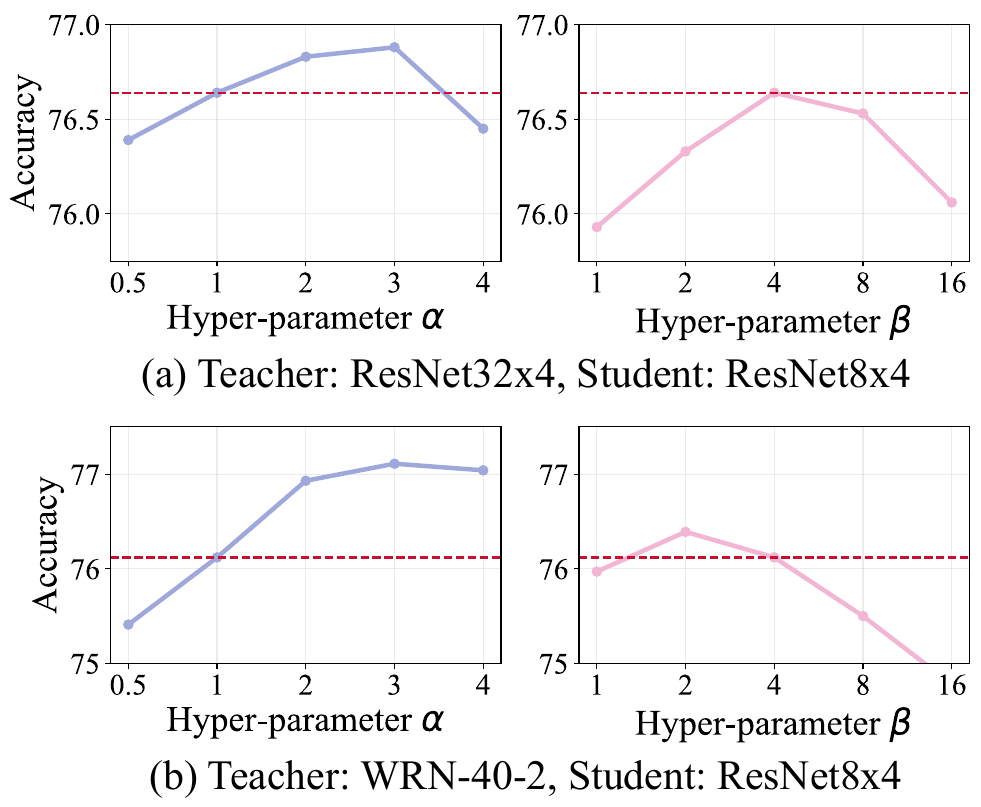}
            \caption{Impact of the hyper-parameters $\alpha$ ($L_\text{SCD}$) and $\beta$ ($L_\text{MCD}$) on the CIFAR-100 validation set. By default, for both distillation pairs, $\alpha=1$ and $\beta=4$ (corresponding to the accuracies reported in \Cref{sec:result}, marked with dashed lines). The reported results are the mean of three trials.}
            \label{fig:parameter}
        \end{figure}

        \paragraph{Hyper-parameter Analysis.} We investigate the impact of the hyper-parameters $\alpha$ and $\beta$, which correspond to the importance of $L_\text{SCD}$ and $L_\text{MCD}$, respectively. As shown in \Cref{fig:parameter}, a more detailed hyper-parameter search can significantly enhance the effectiveness of RLD. Notably, the optimal hyper-parameter configurations vary significantly across different distillation pairs, which fundamentally explains why existing studies~\cite{zhao2022decoupled,sun2024logit} do not adopt fixed hyper-parameter settings. Additionally, we reveal an important phenomenon: unlike DKD, where optimal performance is achieved at $\alpha=1$~\cite{zhao2022decoupled}, RLD generally prefers larger $\alpha$ values. This difference may stem from the following mechanism: when the teacher model makes incorrect predictions, smaller $\alpha$ values in the DKD loss can mitigate the negative impact of incorrect teacher knowledge but also limit the transfer of knowledge to some extent. In contrast, RLD refines the knowledge to effectively eliminate interference from incorrect information, allowing it to adapt to larger $\alpha$ values and further improve model performance.

\section{Conclusion}

    Existing knowledge distillation methods do not consider the impact of incorrect teacher predictions on students. Alternatively, teacher outputs are arbitrarily corrected, disrupting class correlations. In this paper, we introduce Refined Logit Distillation (RLD) to address these issues. RLD enables teacher models to impart two distinct forms of knowledge to the student models: sample confidence and masked correlation. It effectively mitigates over-fitting and eliminates potential misinformation from the teacher models, while maintaining class correlations. Experimental results demonstrate the superiority of RLD.

    \paragraph{Future Work.} There are a few directions to improve our proposed RLD. For instance, dynamic temperature~\cite{li2023curriculum} and meta-learning~\cite{hospedales2021meta} techniques can be used to tune the hyper-parameters. Additional strategies such as data augmentation~\cite{cubuk2019autoaugment} and sample selection~\cite{lan2024improve} can be employed to distill high-quality samples. Besides, combining RLD with state-of-the-art feature distillation methods may be a promising avenue of exploration to further improve the distillation performance. We consider extending correction-based knowledge distillation to the feature domain, utilizing techniques such as Class Activation Mapping~\cite{wang2020score}.

\section*{Acknowledgment}
Wujie Sun and Can Wang are supported by the National Natural Science Foundation of China (No. 62476244), ZJU-China Unicom Digital Security Joint Laboratory and the advanced computing resources provided by the Supercomputing Center of Hangzhou City University.

{
    \small
    \bibliographystyle{ieeenat_fullname}
    \bibliography{main}
}

\appendix

\begin{figure*}[t]
    \centering
    \includegraphics[width=\linewidth]{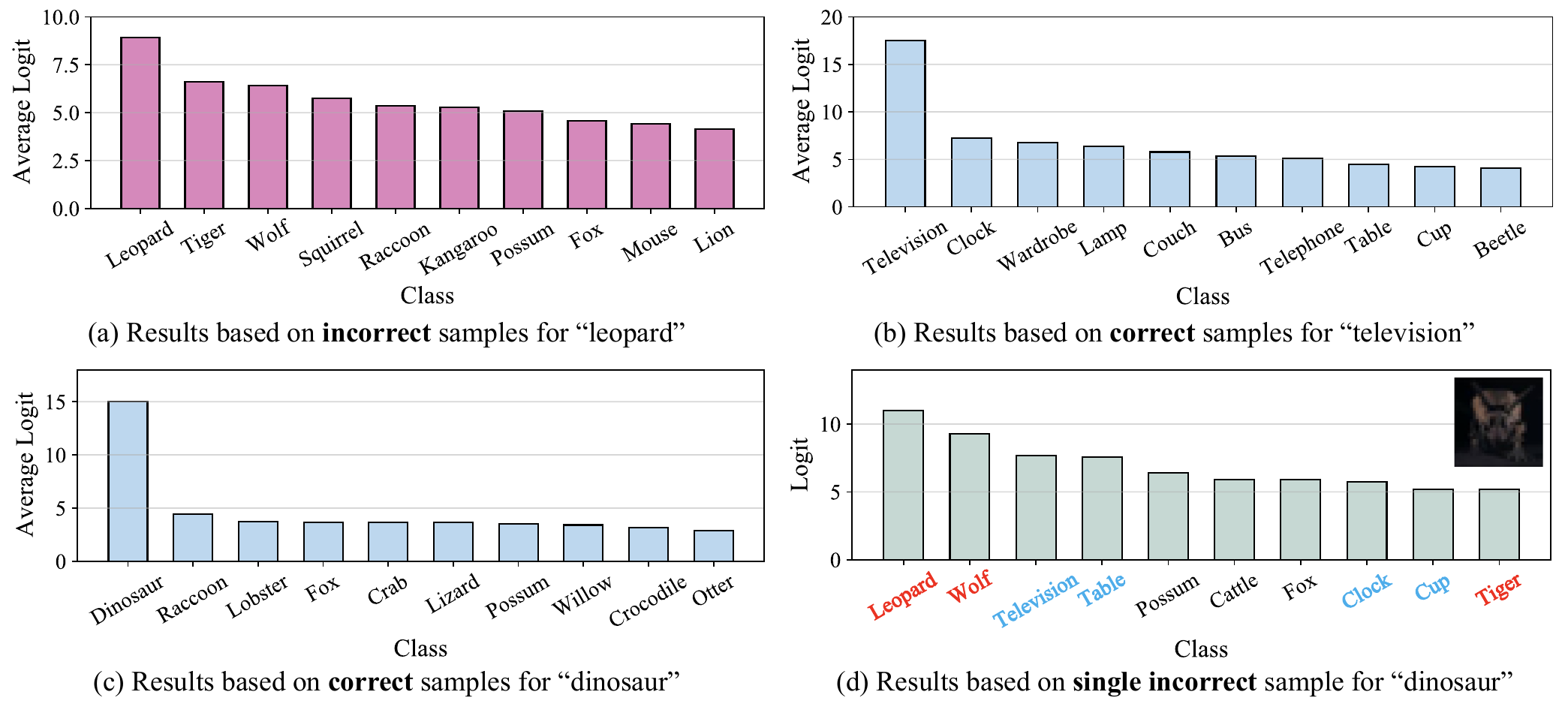}
    \caption{Prediction results for different classes on the CIFAR-100 validation set. The model for calculation is ResNet110. Bold colored classes in (d) denote that they are among the top 10 classes most similar to either ``leopard'' (as inferred from (a)) or ``television'' (as inferred from (b)), yet they do not appear in the top 10 classes most similar to ``dinosaur'' (as inferred from (c)).}
    \label{fig:correlation}
\end{figure*}
    
\section{Class Correlation}

    We present results from a pilot experiment, affirming that scenarios similar to the one depicted in \Cref{fig:correction} do indeed exist. We choose samples specifically from 3 classes in CIFAR-100: leopard, television, and dinosaur. Utilizing a well-trained model, we obtain the top 10 output logits. The corresponding results are displayed in \Cref{fig:correlation}.
    
    \Cref{fig:correlation}(a), (b), and (c) illustrate that the model's class correlations align with our understanding, regardless of whether the prediction is correct or not. For instance, leopard exhibits higher similarity to tiger, while television resembles clock more closely. Across all samples, if the label is ``leopard'', the model does not improve prediction confidence or accuracy by decreasing the logit values of the label's actually similar classes (such as ``tiger'').
    
    \Cref{fig:correlation}(d) reveals that even in the output logits of a single sample, this inter-class relationship is well-maintained. For the sample labeled ``dinosaur'', the top 3 predictions contain ``leopard'' and ``television''. Consequently, within its top 10 predicted classes, the proportion of classes similar to ``leopard'' or ``television'' is significantly high, while classes resembling ``dinosaur'' are rare. In this case, simply correcting the true class (e.g., using augment or swap operations~\cite{cao2023excellent,lan2024improve,wen2021preparing}) can result in significant deviations from the overall class correlations shown in \Cref{fig:correlation}(c), which could negatively impact the distillation performance.
    
\section{Training Efficiency}

\begin{figure}[t]
        \centering
        \includegraphics[width=\linewidth]{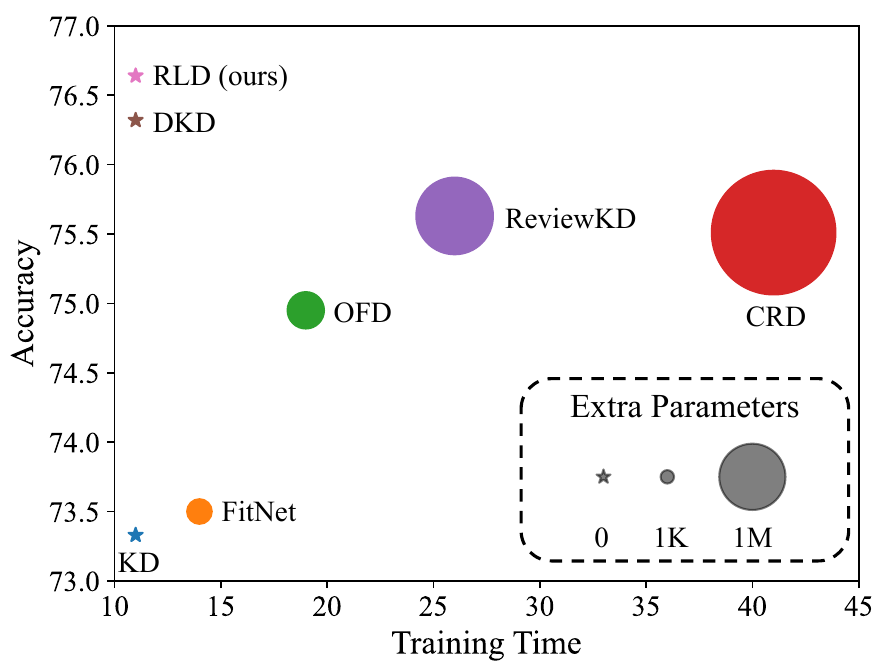}
        \caption{Batch training time (ms) \textit{vs.} top-1 validation accuracy (\%) on the CIFAR-100 dataset. The teacher is ResNet32$\times$4, and the student is ResNet8$\times$4. Larger circle denotes more extra parameters.}
        \label{fig:time}
    \end{figure}

    \Cref{fig:time} presents the training time, accuracy, and extra parameters of various methods. Notably, feature distillation methods (FitNet, OFD, ReviewKD, and CRD) yield a substantial performance enhancement compared to KD, albeit at the cost of a significant increase in training time and extra parameters. In contrast, our RLD method, maintaining the same training time as KD and introducing no extra parameters, delivers superior performance.

\section{Implementation Details}

    We follow the conventional experimental settings of previous works~\cite{tian2020contrastive,zhao2022decoupled,sun2024logit} and use Pytorch~\cite{paszke2019pytorch} for our experiments.
    
    \subsection{CIFAR-100} 
    
        When training on CIFAR-100, the batch size, number of epochs, weight decay, and momentum are set to 64, 240, 5e-4, and 0.9, respectively. The initial learning rates are 0.01 for ShuffleNet and MobileNet, and 0.05 for other model architectures. The learning rate is divided by 10 at 150, 180 and 210 epochs. The optimizer is SGD~\cite{sutskever2013importance}. The training data is augmented using RandomCrop and RandomHorizontalFlip operators.

        For the hyper-parameters involved in RLD, we follow DKD~\cite{zhao2022decoupled} and LSKD~\cite{sun2024logit} to set different values for different distillation pairs. We always set $\alpha$ to 1, and determine the optimal $\beta$ and $\tau$ using grid search from the range of $\{2,4,8,16\}$ and $\{2,3,4,5\}$, respectively.

        Our experiments are carried out on the server equipped with NVIDIA GeForce RTX 2080 Ti GPUs. Each GPU has 11 GB of memory. Only one GPU is used per experiment. The server's operating system is Ubuntu 18.04 LTS.
    
    \subsection{ImageNet} 
    
        When training on ImageNet, the batch size, number of epochs, weight decay, and momentum are set to 512, 100, 1e-4, and 0.9, respectively. The initial learning rate is 0.2 and is divided by 10 every 30 epochs. The optimizer is SGD~\cite{sutskever2013importance}. The training data is augmented using RandomResizedCrop and RandomHorizontalFlip operators.

        For the hyper-parameters involved in RLD, we follow DKD~\cite{zhao2022decoupled} and LSKD~\cite{sun2024logit} to set different values for different distillation pairs. We always set $\alpha$ to 1, and determine the optimal $\beta$ and $\tau$ using grid search from the range of $\{0.5,1,2,3\}$ and $\{1,2\}$, respectively.

        Our experiments are carried out on the server equipped with NVIDIA A800 Tensor Core GPUs. Each GPU has 80 GB of memory. Only one GPU is used per experiment. The server's operating system is Ubuntu 18.04 LTS.

\end{document}